\documentclass{article}

\usepackage{microtype}
\usepackage{graphicx}
\usepackage{subfigure}
\usepackage{booktabs} % for professional tables

\usepackage{hyperref}

\usepackage[accepted]{icml2023}

\usepackage{amsmath}
\usepackage{amssymb}
\usepackage{mathtools}
\usepackage{amsthm}
\usepackage{colortbl}
\usepackage[capitalize,noabbrev]{cleveref}

\theoremstyle{plain}

\theoremstyle{definition}

\theoremstyle{remark}

\usepackage{amsmath,amsfonts,bm}

\def\eqref#1{equation~\ref{#1}}
\def\1{\bm{1}}

\def\vx{{\bm{x}}}
\def\vy{{\bm{y}}}
\def\vz{{\bm{z}}}

\DeclareMathAlphabet{\mathsfit}{\encodingdefault}{\sfdefault}{m}{sl}
\SetMathAlphabet{\mathsfit}{bold}{\encodingdefault}{\sfdefault}{bx}{n}

\def\gC{{\mathcal{C}}}

\def\gF{{\mathcal{F}}}
\def\gG{{\mathcal{G}}}

\def\gJ{{\mathcal{J}}}

\def\gL{{\mathcal{L}}}

\def\gX{{\mathcal{X}}}

\def\gZ{{\mathcal{Z}}}

\def\sR{{\mathbb{R}}}

\newcommand{\E}{\mathbb{E}}

\usepackage[textsize=tiny]{todonotes}

\icmltitlerunning{Latent Traversals in Generative Models as Potential Flows}

\begin{document}

\twocolumn[
\icmltitle{Latent Traversals in Generative Models as Potential Flows}




\begin{icmlauthorlist}
\icmlauthor{Yue Song}{unitn,uva}
\icmlauthor{T. Anderson Keller}{uva,bosch}
\icmlauthor{Nicu Sebe}{unitn}
\icmlauthor{Max Welling}{uva,bosch}
\end{icmlauthorlist}

\icmlaffiliation{unitn}{University of Trento, Italy}
\icmlaffiliation{uva}{University of Amsterdam, the Netherlands}
\icmlaffiliation{bosch}{UvA-Bosch Delta Lab}

\icmlcorrespondingauthor{Yue Song}{yue.song@unitn.it}
%\icmlcorrespondingauthor{Firstname2 Lastname2}{first2.last2@www.uk}

\icmlkeywords{Machine Learning, ICML}

\vskip 0.3in
]



\printAffiliationsAndNotice{}  % leave blank if no need to mention equal contribution

\begin{abstract}
% Abstract (emphasis on latent traversals)
Despite the significant recent progress in deep generative models, the underlying structure of their latent spaces is still poorly understood, thereby making the task of performing semantically meaningful latent traversals an open research challenge. Most prior work has aimed to solve this challenge by modeling latent structures linearly, and finding corresponding linear directions which result in `disentangled' generations. 
In this work, we instead propose to model latent structures with a learned dynamic potential landscape, thereby performing latent traversals as the flow of samples down the landscape's gradient. 
% In this work, we offer a novel perspective on latent structure and associated traversals through the lens of a fluid mechanical interpretation of optimal transport. 
% Specifically, we propose to model the endpoints of a traversal as source and target distributions, and the traversal itself as the gradient flow of a latent variable's probability density over a set of learned potential landscapes. 
Inspired by physics, optimal transport, and neuroscience, these potential landscapes are learned as physically realistic partial differential equations, thereby allowing them to flexibly vary over both space and time. 
To achieve disentanglement, multiple potentials are learned simultaneously, and are constrained by a classifier to be distinct and semantically self-consistent. 
Experimentally, we demonstrate that our method achieves both more qualitatively and quantitatively disentangled trajectories than state-of-the-art baselines. Further, we demonstrate that our method can be integrated as a regularization term during training, thereby acting as an inductive bias towards the learning of structured representations, ultimately improving model likelihood on similarly structured data. Code is available at \url{https://github.com/KingJamesSong/PDETraversal}.
\end{abstract}

\section{Introduction}

Generative models such as Generative Adversarial Networks (GANs)~\citep{goodfellow2014generative} and Variational Auto-Encoders (VAEs)~\citep{kingma2013auto} have latent spaces that are rich in semantics, whereby traversing latent codes according to carefully chosen trajectories has the possibility to lead to semantically meaningful transformations in the generated images. However, without a carefully structured latent space, it is impossible a priori to know how to precisely construct such trajectories. A significant research effort has thus emerged to develop methods that are able to discover semantically meaningful, self-consistent, and disentangled trajectories in the latent space of pre-trained generative models. Such traversals would allow for a more controlled generation of images without needing to alter or constrain the training process of the generative model itself.  Most straightforwardly, an early set of these approaches aimed to identify fixed linear directions in latent space and evolve samples along the discovered directions to create  trajectories~\citep{harkonen2020ganspace,voynov2020unsupervised,shen2021closed}. Such efforts developed valuable techniques for unsupervised learning of interpretable traversal directions but were ultimately limited by their assumption that semantics were structured linearly in latent space, and thus were prone to yielding less semantically disentangled traversals. 
% However, given that the data likely lives on a manifold embedded in a high-dimensional latent space, restricting such traversals to operate linearly is likely to result in a failure to follow true structure of the underlying data, and thus unrealistic traversals. Therefore, a more appropriate traversal trajectory is likely one which is highly non-linear and is able to track the underlying complex manifold. 
More recently, \citet{Tzelepis_2021_ICCV} proposed to model nonlinear latent traversals using gradients of learned Gaussian Radial Basis Functions (RBFs) to effectively `warp' the latent space and thereby drive latent traversals. This integrated non-linearity was demonstrated to improve the modeling of the semantic structure but again was limited by its relatively fixed shape and its static nature over the time-length of the traversal.

% Despite the non-linearity of such traversals being demonstrated useful, their method lacks theoretical support and the corresponding traversal paths of the RBFs are limited in their shape. 
% Except for the linearity setting, existing literature does not pose any other constraints on the properties of traversal paths, such as smoothness or continuity. To sum up, how to formally set up the latent traversal is not well studied yet.

% Andy's Draft
In this work, we introduce a more general framework which encompasses this prior work while simultaneously allowing for a significantly more flexible learned latent structure. Our approach is motivated by intuitions from physics, optimal transport, and neuroscience, and proposes to model latent traversals as the flow of particles down the gradient of a latent potential landscape. The challenge of learning a set of disentangled latent traversals then equates to the problem of learning a set of equivalent disentangled potential functions which match the semantic structure of the underlying data manifold. Traversals can then be generated by evolving samples through time following the gradient of these learned potentials. Importantly, in contrast with prior work, our framework defines the learned potential functions as physically realistic Partial Differential Equations (PDEs), thereby allowing them to vary over both time and space, enabling sufficiently greater flexibility of traversal paths than existing counterparts. In practice, we show that our framework can be applied to multiple different generative models under different experimental settings, and successfully improves performance on a variety of fronts. For example, with pre-trained GANs and VAEs, our framework identifies latent trajectories which are qualitatively more disentangled, and score higher on objective disentanglement metrics than state-of-the-art linear and RBF counterparts. Further, when the desired factors of variation are known a priori, our method can also be integrated into the training process of generative models by performing ``supervised" latent traversals, thereby simultaneously structuring the latent space and providing users with learned latent traversal directions. We show that such integrated structures serve as a beneficial inductive bias for similarly smooth structured input transformations, and thereby improve the likelihood of structured data under the model. Moreover, our latent operator could induce the model with approximate transformation equivarience. Finally, we perform an empirical analysis of our method, demonstrating that our framework can model unambiguous traversal paths in diverse shapes. We conclude with a discussion about how many different well-known `special' PDEs may be used to model the sample evolution, and how previous linear traversal approaches may be seen as special cases of our method.

\section{Motivation}
In this section, we outline the diverse set of motivations which provide useful intuition for the success of our method, in addition to outlining clear paths for potential future work. 

\subsection{Fluid Mechanics as Optimal Transport}
Optimal Transport (OT) can be described at a high level as finding a map which moves the probability mass between a source and target distribution with minimal cost. Intuitively, this has a strong connection with latent traversals which can similarly be seen as attempting to move samples from a source probability distribution to a target probability distribution most efficiently while staying on the data manifold. For example, consider aiming to perform a traversal which changes the length of an individual's hair while leaving the rest of their traits unaffected. With the constraint that the traversal must stay on the data manifold, the most efficient traversal would not involve the transformation of multiple variables, as this would require the movement of additional mass, but instead only transform the latent code in a direction which corresponds to the transformation of a single generative factor. In essence, if we were able to learn the underlying structure of the data manifold with respect to various semantic attributes, optimal transport would give us a direct solution to how to perform disentangled traversals.

One method for solving optimal transport problems involves casting them to a fluid mechanical system  \cite{benamou2000computational}, and solving the associated system numerically. More formally, given the source and target density functions $\rho_0(\vx), \rho_T(\vx) \geq 0$, if we construct a dynamical system defined by a continuous density field $\rho(\vx, t) \geq 0$ and a velocity field $v(\vx, t)$, where $\rho(\vx, 0) = \rho_0(\mathbf{x})$ and $\rho(\vx, T) = \rho_T(\mathbf{x})$, then the classical $L_{2}$ Wasserstein distance can be shown to be equal to the infimum of:
\begin{equation}
\label{eqn:gflow}
    \sqrt{\int_{\mathbf{R}^d} \int_0^T \rho(\vx, t)|v(\vx, t)|^2 \mathop{d\vx dt}}
\end{equation}
over all $v(\vx, t)$ and $\rho(\vx, t)$ which satisfy the continuity equation: $\frac{\partial \rho(\mathbf{x}, t)}{\partial t} = -\nabla\cdot(v(\mathbf{x}, t)\rho(\mathbf{x}, t))$. 
For the individual particles which make up this density field, this corresponds to a time-update in the position given by the vector field at their location, \emph{i.e.}: $\frac{\partial \mathbf{x}}{\partial t} = v(\mathbf{x}, t)$. It turns out that, in terms of the velocity, the optimal solutions to eq.~(\ref{eqn:gflow}) can be written as the gradient of some potential function $\phi$, \emph{i.e.,} $v(\mathbf{x},t)=\nabla_{\mathbf{x}} \phi(\mathbf{x},t)$, thereby earning the name \emph{potential flows}. 
Ultimately, by following such a potential flow, the system can be seen to be minimizing the Wasserstein distance, thereby solving the optimal transport problem.

In relation to latent traversals, we see that we can make an intuitive connection between the distribution of points which make up the start and end points of a given semantic traversal (\emph{e.g.,} the distribution of portraits photos with short and long hair respectively), and the source and target distributions in the OT framework. Following such a connection would intuitively suggest that we may be able to learn a corresponding latent potential $\phi(\vx,t)$ which defines the structure of the latent space with respect to this transformation, and then use the gradient of this field to move particles from one distribution to another. 

% In making a formal connection to latent traversals with generative models, we see there are inconsistencies between the use of probability distributions and the movement of individual particles. 
While making a formal connection with OT remains beyond this paper, we see there is still a close intuitive connection between such methods which may be further formalized in future work. In this work, we present this connection simply as motivation for our method and empirically demonstrate the effectiveness and generality of our approach using this intuition. \textit{One question which comes from this interpretation, is what kind of velocity fields are appropriate for encoding transformations?} In the following subsection, we provide further intuition that motivates our use of physically-realistic PDEs such as the wave equation to constrain the space-time dynamics of $\phi$ and the resulting velocity $\nabla\phi$.

% the infimum of the classical $L_{2}$ Wasserstein distance can be shown to be equivalent to the infimum of an equivalent 

% \begin{theorem}[$L_2$ Kantorovich Problem]
% For probability measures $\mu_{0}$ and $\mu_{1}$, the $L_{2}$ Wasserstein distance can be defined as
% \begin{equation}
%     W_2(\mu_{0},\mu_{1}) = \min \{\sqrt{\int\int\rho(x,t)|v(x,t)|^2\mathop{dx}\mathop{dt}} \}
% \end{equation}
% where the density $\rho$ and the velocity $v$ satisfy:
% \begin{equation}
%     \frac{\mathop{d}\rho(x,t)}{\mathop{dt}} = -\nabla\cdot(v(x,t)\rho(x,t)),\ v(x,t)=\nabla\phi(x,t)
% \end{equation}
% and the potential $\phi$ satisfies the Hamilton-Jacobi equation:
% \begin{equation}
%      \frac{\mathop{d}\phi(x,t)}{\mathop{dt}} + \frac{1}{2}|\nabla\phi(x,t)|^2 =0
% \end{equation}
% \end{theorem}

% \paragraph{Continuum Mechanics} 

\subsection{Traveling Waves in Neuroscience} 
More abstractly, our work is motivated by the recent interest in traveling waves in the neuroscience literature. Succinctly, traveling waves have recently been observed to exist in a diversity of regions and scales in the biological cortex~\citep{muller2018cortical}. Although a consensus has yet to be reached about their exact computational purpose, there is a variety of emerging work which appears to implicate them in the predictive processing of observed transformations from both biological \citep{jancke2004imaging, sato2012traveling, friston2019waves, alamia2019alpha, besserve2015shifts} and computational \cite{cosyne-waves} perspectives. Specifically, these works suggest that they play the role of integrating information across time, encoding motion, and modulating information transfer. In this work, we leverage these observations to motivate the hypothesis that \textit{traveling waves may be a neural correlate of latent traversals, and thereby serve as an efficient way to encode natural transformations using neural network architectures.} Pursuant to this hypothesis, we expect beneficial performance with physics-inspired PDEs guiding latent traversals in artificial neural networks as well.
 % thus modeling latent structure in this manner and
% Research suggests that they may serve a role in integrating information across time~\citep{sato2012traveling}, encoding motion~\citep{jancke2004imaging}, modulating information transfer~\citep{besserve2015shifts}, and facilitating predictive coding~\citep{friston2019waves,alamia2019alpha}, but a formal consensus on their computational function has yet to be reached. In this work, we regulate our 
\section{Related Work}

\noindent\textbf{Latent Traversal in Generative Models.} Latent traversals have often been used to evaluate the quality of learned latent spaces of the deep generative models \citep{kingma2013auto, goodfellow2014generative}. Pursuant to this, much research has been conducted to determine the optimal way to compute traversal trajectories in order to yield semantically meaningful generations. 
% Generative models such as VAEs~\citep{kingma2013auto} and GANs~\citep{goodfellow2014generative} are known to possess rich semantics in the structured latent space. Moving around the latent codes in specific interpretable vector directions could trigger semantically-meaningful variations in the output images~\citep{radford2015unsupervised}. 
One line of research employs explicit human annotations to define the semantic labels for interpretable paths~\citep{radford2015unsupervised,goetschalckx2019ganalyze,jahanian2020steerability,plumerault2020controlling,shen2020interpreting,ling2021editgan,shi2022semanticstylegan}. By contrast, unsupervised methods discover interpretable directions without any prior knowledge~\citep{harkonen2020ganspace,kwon2022diffusion,choi2021not,karmali2022hierarchical,spingarn2020gan,ren2021learning,oldfield2022panda}. For example, \citet{voynov2020unsupervised} proposed to learn a set of semantic concepts via an auxiliary classifier. Other methods such as SeFa~\citep{shen2021closed} pointed out that the eigenvectors of the projection matrix following the latent codes can be directly used as interpretable directions. More recently, \citet{Tzelepis_2021_ICCV} proposed to non-linearly perturb the latent code using gradients of learned RBFs. Our work mainly belongs to the unsupervised category, as demonstrated by the majority of the results presented in Sec.~\ref{sec:experiments}; however, as we show in Sec.~\ref{sec:supervised_method} and \ref{sec:supervised_exp}, our method can also be extended to the supervised setting, thereby regularizing the latent space towards increased structure and improving the model's ability to represent similarly structured transformations.

%The most simple way is to change the latent code in a single dimension. However, this may not exploit the latent structure and fail to. More recently, \citet{Tzelepis_2021_ICCV} proposed to perturb the latent code non-linearly. 

\noindent\textbf{Disentanglement Learning.} In contrast to the goal of discovering latent traversal trajectories in pre-trained models, other methods have aimed to attain an a priori structured representation through additional regularization during training. For example, InfoGAN~\citep{chen2016infogan} encouraged disentanglement by maximizing the mutual information between the observations and a fixed subset of the latent code. \citet{zhu2020learning} proposed a variational predictability loss to learn disentangled representations and introduced a metric to evaluate unsupervised disentanglement methods. \citet{peebles2020hessian,wei2021orthogonal} and \citet{song2022orthogonal} proposed different orthogonality constraints to improve disentanglement ability. Alternatively, for disentanglement with VAEs, much work has focused on various modifications to the evidence lower bound (ELBO) to encourage increased independence of the different latent dimensions. Most notably, the $\beta$-VAE~\citep{higgins2016beta} first introduced a hyper-parameter to accentuate the penalty of the divergence between the prior and variational posterior. Follow-up research used additional guidance to encourage improved disentanglement in this manner, including $\beta$-TC-VAE~\citep{kim2018disentangling,chen2018isolating}, DIP-VAE~\citep{kumar2018variational}, Guided-VAE~\citep{ding2020guided}, JointVAE~\citep{dupont2018learning}, and CasadedVAE~\citep{jeong2019learning}.

%Traversal is such an approach to learn more disentangled representations, which we will discuss below.

%\noindent\textbf{Travelling heats.} From the literature of neuroscience, travelling heats are known to exist at diverse regions and scales in biological cortex~\citep{muller2018cortical}. Some research suggest or hypothesize that they may serve the role of integrating information across time~\citep{sato2012traveling}, encoding motion~\citep{jancke2004imaging}, modulating information transfer~\citep{besserve2015shifts}, and facilitating predictive coding~\citep{friston2019heats,alamia2019alpha}. Also, the heat PDEs satisfy the continuity equation, \emph{i.e.,} the mass is conserved during the travelling. 
%Our work is inspired by their connection to generative modelling: the ideal latent traversal of image attributes (\emph{e.g.,} head pose) should have only one variation factor and other contents should be preserved, which may loosely correspond to the conservation of mass and the role of travelling heats in neuroscience.

%\noindent\textbf{Physics Informed Neural Networks.} 

\noindent\textbf{Physics for Deep Learning.} In recent years, an increased effort has developed to combine deep neural networks with concepts from physics. Much work has focused on using deep learning to solve problems that arise in physics, such as solving PDEs by Physics Informed Neural Networks (PINNs)~\cite{pinns}, learning dynamic systems with Neural ODEs~\cite{chen2018neural}, and discovering physical concepts~\cite{iten2020discovering}. Another active research field leverages fundamental laws (\emph{e.g.,} symmetries or conservation laws) to improve deep learning models. Some examples include designing equivariant neural networks to handle input with geometric symmetries~\cite{cohen2016group,cohen2018spherical,zhang2019making,satorras2021n,keller2021topographic}, endowing neural networks with Hamiltonian dynamics for improved performance and generalization~\cite{greydanus2019hamiltonian,toth2019hamiltonian}, and building score-based denoising diffusion models for generative modelling~\cite{ho2020denoising,song2020denoising,song2020score}. In this work, we use PINN-inspired constraints to model the latent traversal with learned potential PDEs, situating our model in the category of work which seeks to improve deep learning with physically inspired methods.

%\noindent\textbf{Optimal Transport in Deep Learning.} There is a vast literature of OT theory and applications in various fields~\citep{villani2009optimal,villani2021topics}. Here we mainly highlight on the relevant applications in machine learning. The pioneering work of \citet{cuturi2013sinkhorn} proposed a lightspeed implementation of Sinkhorn algorithm for fast computation of entropy-regularized Wasserstein distance, which opens the way for many differentiable Sinkhorn algorithm based applications~\citep{frogner2015learning,feydy2019interpolating,chizat2020faster,eisenberger2022unified,kolouri2020wasserstein}. Specific in generative modeling, the Wasserstein distance is often used to minimize the discrepancy between the data distribution and the model distribution~\citep{arjovsky2017wasserstein,tolstikhin2017wasserstein,salimans2018improving,patrini2020sinkhorn}. Notice that most previous works compute the OT plan using the Sinkhorn operators or directly penalize the Wasserstein distance. Our framework is based on the fluid mechanics interpretation of OT, which is rarely attempted in the field of deep learning.

\begin{figure*}[t]
    \centering
    \includegraphics[width=0.99\linewidth]{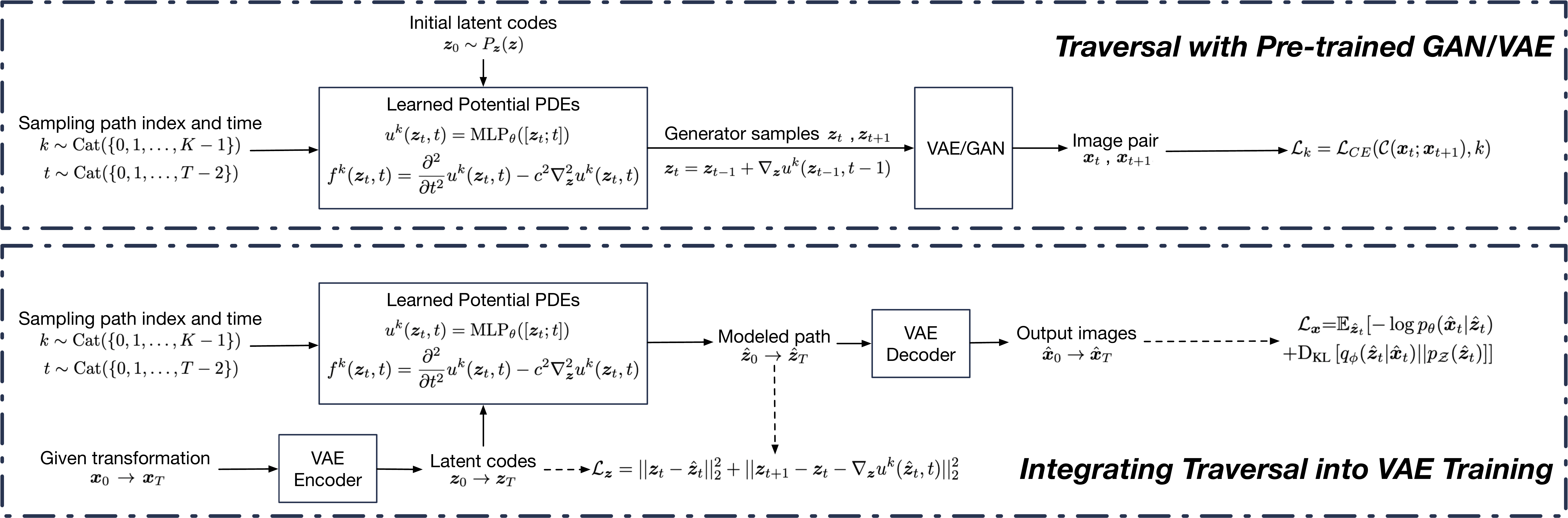}
    \caption{Overview of our learned potential PDEs for latent traversal in two different experimental settings.}
    \label{fig:workflow}
\end{figure*}

\section{Methodology}

%This section presents our novel perspective of considering latent traversal as OT problems, the connection of OT with fluid mechanics, the formulation of our learned PDEs, the training strategies, and how our method is integrated into GANs and VAEs under different experimental settings.

In this section we present the formulation of our learned potential functions, their integration into generative models under different settings, and the training and sampling strategies. The overview of our method is depicted in Fig.~\ref{fig:workflow}.

\subsection{Latent Traversals as Potential Flows}
\label{sec:method_pde}

\noindent\textbf{Learning the Potential PDE.} Assume we are given a pre-trained generative model $\gG: \gZ \rightarrow \gX$ with prior distribution $P_{\vz}(\vz)$. To model $K$ different semantically disentangled latent trajectories, we model each trajectory separately as the gradient of a learned time-dependant scalar potential energy field: $u^k(\vz_{t}, t) = \mathrm{MLP}_{\theta^k}([\vz_{t}; t]) \ \in \sR$. In this work we use a small multilayer perceptron (MLPs) to learn each potential. The process of traversing from an initial sample ($\vz_0$) to a future element ($\vz_t$) at time $t$ is then defined as the potential flow $\nabla_\vz u$ described by this field:
\begin{equation}
\begin{split}
\vz_0 \sim P_{\vz}(\vz)  \ \ \ \ \ \ \vz_{t} = \vz_{t-1} + \nabla_\vz u^k(\vz_{t-1}, t-1)
\end{split}
\end{equation}
To encourage the latent potential to model realistic trajectories and follow the intuitions outlined above, we additionally impose a PINN constraint in the form of the second-order wave equation with wave coefficient $c$: 
\begin{equation}
\begin{split}
f^k(\vz_{t}, t) &= \frac{\partial^2}{\partial t^2} u^k(\vz_{t},t) 
 - c^2 \nabla^2_\vz u^k(\vz_{t},t)
\end{split}
\end{equation}
Such a constraint makes our potential flow model a good approximation of small amplitude sound waves \cite{Lamb1993-wl}, and empirically is seen to produce highly diverse and realistic trajectories. Our objective is then to minimize:
\begin{equation}
\begin{split}
\gL_f = \frac{1}{T}\sum^{T-1}_{t=0} ||f^k(\vz_t, t)||_2^2,\  
\gL_u = ||\nabla_\vz u^k(\vz_0, 0)||_2^2 \\
\end{split}
\end{equation}
where $T$ represents the total number of timesteps of our latent trajectory, $\gL_f$ restricts the energy to obey our physical constraints, and $\gL_u$ restricts $u(\vz_t, t)$ to return no update at $t{=}0$, thereby matching the initial condition.

\noindent\textbf{Jacobian Regularization.} While the above formulation models traversals as physically realistic potential flows, it cannot ensure that the modeled traversal paths are semantically meaningful. Therefore, to make our learned potentials more aligned with the semantics of the data, we take inspiration from prior work and further couple the traversal direction with the Jacobian of the generator. Similar to~\citet{zhu2021low,zhu2022region}, we first approximate the manipulation on the latent space as
\begin{equation}
    \gG(\vz_{t} + \epsilon\nabla u^k(\vz_{t}, t))\approx\gG(\vz_{t})+ \epsilon\underline{\frac{\partial \gG(\vz_{t})}{\partial \vz_{t}} \nabla_{\vz} u^k(\vz_t, t)}
    \label{eq:jvp}
\end{equation}
where $\epsilon$ denotes perturbation strength. Intuitively, for sufficiently small $\epsilon$, if the Jacobian-vector product (the underlined term in eq.~(\ref{eq:jvp})) can cause large variations in the generated sample, the direction is likely to be semantically meaningful. We therefore introduce a Jacobian-vector product regularization term to encourage the improved semantic variations of our traversals in an unsupervised manner: 
\begin{equation}
    \gL_{\gJ} =-||\frac{\partial \gG(\vz_{t})}{\partial \vz_{t}} \nabla_\vz u^k(\vz_{t}, t)||_2^2
\end{equation}

%subsection{Training Strategies}

%Then the generator is fed with two latent codes $\vz_0$ and $\vz_0 + \nabla_z u^k(\vz_0, t)$, and outputs two different images. Finally, the auxiliary discriminator takes as input the images and predicts the trajectory index $\hat{k}$ and the timestamp $\hat{t}$. depending on the specific generative model (GAN or VAE), to encourage that each heat corresponds to different semantics.

\subsection{Traversal with Pre-trained GAN/VAE}
With pre-trained models, the weights of the generator are frozen. We only update the parameters of our MLPs and of the auxiliary potential-index classifier module. We adopt an auxiliary classifier $\gC$ to predict the potential index and use the cross-entropy loss to optimize it:
\begin{equation}
    \hat{k}{=}\gC(\vx_{t};\vx_{t+1}),\ \gL_{k}=\gL_{CE}(\hat{k},k)
\end{equation}
Where $\vx_{t} = \gG(\vz_t)$ is the generated sample from timestep $t$.

%\noindent\textbf{GANs.} 

%\noindent\textbf{VAEs.} In the VAE setting, we learn the approximate posterior $q_{\gamma}(k | \vx_t, \vx_{t+1})$ to infer the trajectory index as
%\begin{equation}
%\begin{split}
    %\label{eq:vae_elbo}
    %\gL_{k}=\mathrm{D}_{\text{KL}}\left[q_{\gamma}(k | \vx_{t}, \vx_{t+1}) %|| \text{Cat}(k)\right] \\+ \gL_{CE}(p(k|\vx_{t}, \vx_{t+1}),k)
 %   \gL_{k}=\gL_{CE}(\hat{k},k)
%\end{split}
%\end{equation}
%where $\mathrm{D}_{\text{KL}}$ denotes the KL divergence measure. 

\begin{figure*}[t]
    \centering
    \includegraphics[width=0.99\linewidth]{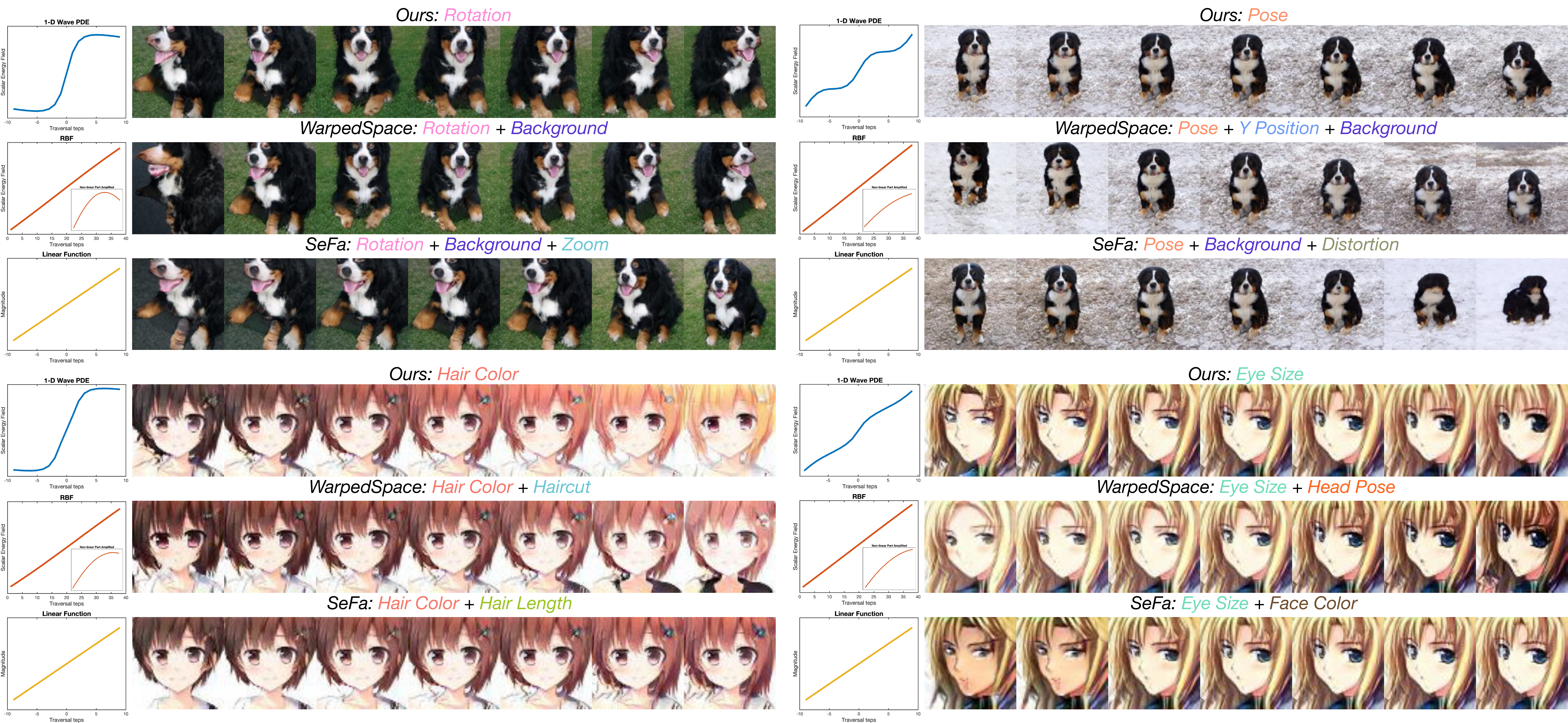}
    \caption{Exemplary traversal paths (potential PDEs for our method) and the corresponding interpolation images with SNGAN and BigGAN. Since the paths of WarpedSpace are of very limited non-linearity that is hard to perceive, we amplify the non-linear part in the sub-figure inside the figure as follows: for a traversal path $\vy$ of WarpedSpace, we decompose it into $\vy = \vy_{LN} + \vy_{NLN}$ where $\vy_{LN}$ denotes the linear part and $\vy_{NLN}$ is the non-linear counterpart. Then the non-linearity part is amplified by $\vy = \vy_{LN} + 200\cdot\vy_{NLN}$.}
    \label{fig:quali_anime_dog}
\end{figure*}

\subsection{Integrating Traversal into VAE Training}
\label{sec:supervised_method}
When training VAEs from scratch, our method can perform ``supervised" latent traversal as extra regularization to improve the likelihood. That is, we explicitly model the path of the variations of a semantic attribute during the training process. In this setting, we consider having access to the pre-defined transformation of each variation factor $\vx_0 \rightarrow \vx_T$. Then we can obtain the corresponding latent codes $\vz_0 \rightarrow \vz_T$ by feeding images to the encoder, \emph{i.e.,} $\vz_{t}=\texttt{Encode}(\vx_{t})$. Then our potential PDEs manipulate the initial latent codes $\vz_{0}$ to obtain $\hat{\vz}_1 \rightarrow \hat{\vz}_T$ by progressively performing $\hat{\vz}_{t}=\vz_{0}+\sum\nabla_\vz u^k$. The output images $\hat{\vx}_1 \rightarrow \hat{\vx}_T$ can be easily attained by decoding $\hat{\vz}_1 \rightarrow \hat{\vz}_T$. The traversal paths modeled by our wave equations are encouraged to match the ground truth as
\begin{equation}
\begin{aligned}
 \gL_{\vz} &= ||\vz_{t}-\hat{\vz}_{t}||_{2}^{2} + ||(\vz_{t+1}-\vz_{t}) -(\hat{\vz}_{t+1}-\hat{\vz}_{t}) ||_{2}^{2} \\
 &= ||\vz_{t}-\hat{\vz}_{t}||_{2}^{2} + ||\vz_{t+1}-\vz_{t}-\nabla_\vz u^k(\hat{\vz}_{t}, t)||_{2}^{2}
\end{aligned}
\end{equation}
where the first term penalizes the difference between current latent codes and the ground truth history, and the second term ensures that the future update at the next timestep is realistic. Besides improving the plausibility of traversal paths, we optimize the ELBO:
\begin{equation}
\begin{split}
    \gL_{\vx} {=} \E_{\hat{\vz}_t}[ {-}\log p_{\theta}(\hat{\vx}_t | \hat{\vz}_t)  {+} \mathrm{D}_{\text{KL}}\left[q_{\phi}(\hat{\vz}_t | \hat{\vx}_t) || p_{\gZ}(\hat{\vz}_t)] \right ]
    %-\log p_{\theta}(\hat{\vx}_0 | \hat{\vz}_0)
\end{split}
\end{equation}
where $p_{\theta}$ parameterizes the generator, and $q_{\phi}$ denotes the approximate posterior. The combination of the two losses could yield more structured latent space and more realistic traversal trajectories, which might improve the likelihood. 

%and the output images $\hat{\vx}_0 \rightarrow \hat{\vx}_T$

%We also use eq.~(\ref{eq:ce}) to learn to discriminate the trajectories. Notice that we do not include eq.~(\ref{eq:vae_elbo}) in the above ELBO because we consider regularizing the model of individual samples instead of the model of sequences.

%The overall loss function is:
%\begin{equation}
%    \gL = \gL_u + \gL_f + \gL_{\gJ} + \gL_{k} + \gL_{\vz} + \gL_{\vx} 
%\end{equation}

%and also the latent codes $\hat{\vz}_0 \rightarrow \hat{\vz}_t$ predicted by our heat PDEs. We still sample two consecutive steps $t$ and $t-1$, and the overall loss function is defined as:

\subsection{Sampling and Training Strategies} 

At each training step, we randomly sample a potential index $k$ from $\mathrm{Cat}(\{0, 1, \ldots, K{-}1\})$ and a timestep $t$ from $\mathrm{Cat}(\{0, 1, \ldots, T{-}2\})$. Then we use the selected potential to generate the corresponding velocity fields and obtain the two latent codes $\vz_{t}$ and $\vz_{t+1}$. Subsequently, the generator is fed with the latent codes and outputs a pair of images $\vx_{t}$ and $\vx_{t+1}$. Finally, we adopt an auxiliary classifier to predict the potential index $\hat{k}$. The overall loss function is defined as
\begin{equation}
\label{eq:ce}
    \gL = \gL_u + \gL_f + \underline{\gL_{\gJ} + \gL_{k}} + \boxed{\gL_{\vx} + \gL_{\vz}}\\
\end{equation}
where $\gL_{k}$ matches the predicted index $\hat{k}$ to the ground truth $k$, therefore encouraging that each learned potential is significantly distinct and self-consistent to be recognized by a classifier accurately. The boxed terms are only applied to regularize the latent space when integrated into VAE training, while the underlined terms are used for pre-trained models. Notice that different from~\citet{voynov2020unsupervised,Tzelepis_2021_ICCV}, we do not predict the timesteps from the image pair $[\vx_{t},\vx_{t+1}]$. This is because our potential PDEs can be very diverse in spatiotemporal form, thus predicting the timesteps from two points on the path demonstrated to be both unnecessary and practically infeasible.

%The implementation of $\gL_{k}$ depends on specific generative models (GANs or VAEs), which we would introduce later. 

\begin{figure*}[tbp]
    \centering
    \includegraphics[width=0.99\linewidth]{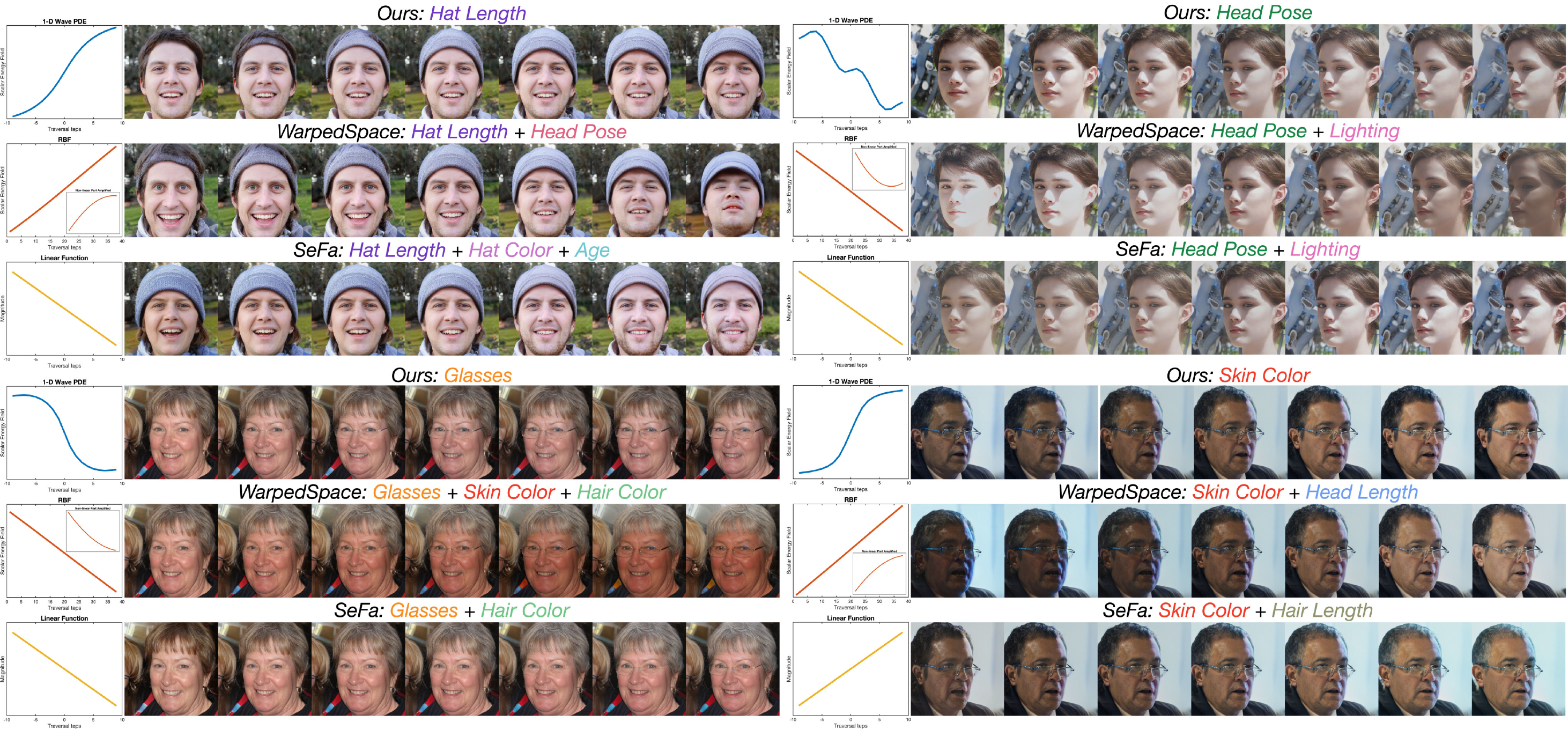}
    \caption{Traversal trajectories (potential PDEs for our method) and the associated interpolation images of the exemplary four attributes with StyleGAN2. The non-linearity of WarpedSpace paths is amplified in the same way as done in SNGAN and BigGAN.}
    \label{fig:quali_stylegan2}
\end{figure*}

\section{Experiments}
\label{sec:experiments}
This section starts with the setup, followed by the results under different settings, and ends with in-depth discussions.

\subsection{Settings} 

\noindent\textbf{Models and Datasets.} For experiments of pre-trained GANs, our method is evaluated on SNGAN~\citep{miyato2018spectral} with AnimeFace~\citep{chao2019/online}, BigGAN~\citep{brock2019large} with ImageNet~\citep{deng2009imagenet}, and StyleGAN2~\citep{karras2020analyzing} with FFHQ~\citep{karras2019style}. For BigGAN, we train the target class ``Bernese mountain dog". We adopt LeNet~\citep{lecun1998gradient} as the auxiliary classifier for SNGAN, while ResNet-18~\citep{he2016deep} based classifier is used for both BigGAN and StyleGAN2. For the VAEs experiments, we use the VAE encoder as the auxiliary classifier and evaluate our method on MNIST~\citep{lecun1998mnist} and dSprites~\citep{dsprites17} datasets.

\noindent\textbf{MLP for Modeling PDEs.} We use sinusoidal positional embeddings~\cite{vaswani2017attention} to embed the timestep $t$. Linear layers with \texttt{Tanh} activations are used for embedding the latent code input $\vz$. Another linear layer is used to fuse features across space and time. We set the wave coefficient $c$ as a learnable parameter and initialize it with $1$.

%Finally, we also use \texttt{Tanh} to limit the output range.
%
\noindent\textbf{Metrics.} For the quantitative evaluation of traversal with GANs, we use Variational Predictability (VP)~\cite{zhu2020learning} score and the correlation coefficient between face attributes and traversal steps using pre-trained attribute estimators. The VP score adopts the few-shot learning setting (\emph{e.g.,} 10\% images as the training set) to measure the generalization of a simple neural network in classifying the discovered latent directions from a crafted dataset of random image pairs $[\vx_{0},\vx_{T}]$. For attribute correlation, we first use S3FD~\citep{zhang2017s3fd} to extract the face region and then compute the normalized Pearson’s correlation between potential indexes and traversal steps using several pre-trained attributes estimators, including ArcFace~\citep{deng2019arcface} for face identity, FairFace~\citep{karkkainen2021fairface} for face attributes (age, race, and gender), and HopeNet~\citep{doosti2020hope} for face poses (yaw, pitch, and roll). The correlation results are averaged across $50$ random latent samples. For the quantitative evaluation of VAEs, since our method performs vector-based manipulation, traditional single-dimension-based VAE disentanglement metrics such as Mutual Information Gap (MIP)~\citep{chen2018isolating} do not apply here. Some works such as~\citet{arvanitidis2017latent,tonnaer2020quantifying} can perform the evaluation of quantitative vector-based manipulation but they require supervision of the ground truth. We thus also evaluate the disentanglement performance using the VP score. The log-likelihood over the entire dataset is measured for the experiment of integrating our method into  the VAE training.

%In line with~\citet{Tzelepis_2021_ICCV}, 

\noindent\textbf{Baselines.} For pre-trained GANs, we compare our method against two representative baselines, \emph{i.e.,} SeFa~\citep{shen2021closed} and WarpedSpace~\citep{Tzelepis_2021_ICCV}. SeFa uses eigenvectors of the weight matrix after latent codes for \textit{linear} perturbation, while WarpedSpace \textit{non-linearly} changes the latent codes using the gradients of RBFs. As for VAEs, there are no popular vector-based traversal methods in the literature so we also use WarpedSpace for comparison. Finally, as another controlled baseline, we train a linear function with other settings aligned with our method. 
%train a linear function with other settings unchanged as a controlled baseline.

%our method is compared against SeFa and naively-trained baselines. Finally, as another controlled baseline, we also train a linear function with other settings unchanged. 
\begin{table}[h]
    \centering
    \caption{Comparison of the VP scores (\%) with different GANs. The results are averaged over $3$ random runs.}
    %\resizebox{0.79\linewidth}{!}{
    \begin{tabular}{c|c|c|>{\columncolor[gray]{0.8}}c}
    \toprule
         \rowcolor{white}\textbf{Models} & \textbf{SeFa} & \textbf{WarpedSpace} & \textbf{Ours}\\
    \midrule
          \textbf{SNGAN}  &53.76 &58.83  &\textbf{65.89} \\
          \textbf{BigGAN} &13.59 &14.07  &\textbf{15.29} \\
          \textbf{StyleGAN2} &39.20 &36.31  & \textbf{48.54}\\
    \bottomrule
    \end{tabular}
    %}
    \label{tab:vp_score}
\end{table}

%\noindent\textbf{Inference.}

\begin{table}[t]
    \centering
     \caption{The $l_{1}$ normalized attribute correlations of our method (\emph{top}), WarpedSpace (\emph{middle}), and SeFa (\emph{bottom}) based on $50$ samples. The second highest correlation is also highlighted if the best value in the row is not on the diagonal.}
    \resizebox{0.99\linewidth}{!}{
    \begin{tabular}{c|c|c|c|c|c|c|c}
    \toprule
         & Yaw & Pitch & Roll & Identity &Age & Race & Gender  \\
    \midrule
      Yaw &\cellcolor{gray!50}{\textbf{0.34}} &0.09 &0.22 &0.09 &0.03 &0.18 & 0.03 \\
      Pitch &0.04 &\cellcolor{gray!20}{\textbf{0.25}} &0.11 &0.08 &0.00 &0.08 &\cellcolor{gray!50}{\textbf{0.45}} \\
      Roll &0.23 &0.19 &\cellcolor{gray!50}{\textbf{0.35}} &0.00 &0.02 &0.03 &0.18\\
      Identity &0.01 &0.06 &0.00 &\cellcolor{gray!50}{\textbf{0.61}} &0.21 &0.03 & 0.07 \\
      Age & 0.00 &0.06 &0.00 &0.03 &\cellcolor{gray!50}{\textbf{0.87}} &0.00 &0.04 \\
      Race &0.05 &0.07 &0.06 &0.02 &0.01 &\cellcolor{gray!50}{\textbf{0.73}} &0.06 \\
      Gender &0.08 &0.19 &0.09 &0.04 &0.00 &0.03 &\cellcolor{gray!50}{\textbf{0.58}} \\
    \bottomrule
    \end{tabular}
    }
    \resizebox{0.99\linewidth}{!}{
    \begin{tabular}{c|c|c|c|c|c|c|c}
    \toprule
         & Yaw & Pitch & Roll & Identity &Age & Race & Gender  \\
    \midrule
      Yaw &\cellcolor{gray!20}{\textbf{0.34}} &0.03 &0.05 &\cellcolor{gray!50}{\textbf{0.42}} &0.01 &0.08 &0.07  \\
      Pitch &0.01 &\cellcolor{gray!20}{\textbf{0.38}} &0.07 &\cellcolor{gray!50}{\textbf{0.42}} &0.01 &0.09 &0.01 \\
      Roll &0.10 &0.15 &0.17 &\cellcolor{gray!50}{\textbf{0.27}} &0.02 &0.07 &\cellcolor{gray!20}{\textbf{0.22}} \\
      Identity &0.01 &0.10 &0.00 &\cellcolor{gray!50}{\textbf{0.69}} &0.10 &0.07 &0.01  \\
      Age &0.02 &0.09 &0.05 &\cellcolor{gray!50}{\textbf{0.52}} &\cellcolor{gray!20}{\textbf{0.25}} &0.02 &0.05 \\
      Race &0.05 &0.02 &0.07 &0.12 &0.07 &\cellcolor{gray!50}{\textbf{0.54}} &0.12 \\
      Gender &0.09 &0.00 &0.02 &0.40 &0.00 &0.00 &\cellcolor{gray!50}{\textbf{0.49}} \\
    \bottomrule
    \end{tabular}
    }
    \resizebox{0.99\linewidth}{!}{
    \begin{tabular}{c|c|c|c|c|c|c|c}
    \toprule
         & Yaw & Pitch & Roll & Identity &Age & Race & Gender  \\
    \midrule
      Yaw &\cellcolor{gray!20}{\textbf{0.29}} &0.01 &0.05 &\cellcolor{gray!50}{\textbf{0.40}} &0.04 &0.09 &0.11 \\
      Pitch &0.09 &\cellcolor{gray!20}{\textbf{0.29}} &0.06 &\cellcolor{gray!50}{\textbf{0.41}} &0.05 &0.08 &0.01 \\
      Roll &0.03 &0.10 &0.09 &\cellcolor{gray!50}{\textbf{0.60}} &0.00 &0.06 &\cellcolor{gray!20}{\textbf{0.12}} \\
      Identity &0.02 &0.05 &0.02 &\cellcolor{gray!50}{\textbf{0.74}} &0.08 &0.08 &0.01  \\
      Age &0.02 &0.08 &0.02 &\cellcolor{gray!50}{\textbf{0.47}} &\cellcolor{gray!20}{\textbf{0.25}} &0.02 &0.15 \\
      Race &0.07 &\cellcolor{gray!20}{\textbf{0.25}} &0.02 &\cellcolor{gray!50}{\textbf{0.58}} &0.00 &0.00 &0.07 \\
      Gender &0.02 &0.05 &0.02 &\cellcolor{gray!50}{\textbf{0.43}} &0.02 &\cellcolor{gray!20}{\textbf{0.35}} &0.12\\
    \bottomrule
    \end{tabular}
    }
    \label{tab:att_corr}
\end{table}

\subsection{Results with Pre-trained GANs}

\noindent\textbf{SNGAN and BigGAN.} Fig.~\ref{fig:quali_anime_dog} displays the exemplary latent traversal results and the corresponding trajectories with SNGAN and BigGAN. Since the parameters of the generator are frozen, each method would generate the same image for one latent sample. Our PDEs can generate traversal paths with distinct semantics and precise image attribute control, while the baselines suffer from entangled attributes and the non-target semantics also vary during traversal. Moreover, the paths of WarpedSpace are of very limited non-linearity, which is imperceptible unless the non-linear part of the path is significantly amplified. By contrast, our potential PDEs have more diverse shapes and more flexible non-linearity. Table~\ref{tab:vp_score} presents the quantitative evaluation results of the VP scores. Our PDEs achieve state-of-the-art performance in terms of classification accuracy in the few-shot learning setting. Specifically, our method outperforms the second-best baseline by $7.04\%$ with SNGAN, by $1.22\%$ with BigGAN, and by $12.23\%$ with StyleGAN2. The consistent performance gain on each dataset indicates that the semantics of our traversal paths are indeed more disentangled than others. It is also worth mentioning that the relatively marginal advantage with BigGAN might stem from the fact that BigGAN generates images in wide domains ($1,000$ ImageNet classes). This domain diversity might restrict the actual number of latent semantics, thus limiting the performance.

%The highest correlation is highlighted in \cellcolor{gray!50}{\textbf{blue}}, and the second highest is highlighted in \cellcolor{gray!20}{\textbf{cyan}} if it is not on the diagonal.

\noindent\textbf{StyleGAN2.} Fig.~\ref{fig:quali_stylegan2} compares the exemplary latent traversal with StyleGAN2. The results are coherent with those on SNGAN and BigGAN: the traversal paths of baselines suffer from entangled semantics, while our potential PDEs are able to model trajectories that correspond to more disentangled image attributes. Table~\ref{tab:att_corr} presents the $l_{1}$ normalized correlation results of some common face attributes. As can be seen, most attributes of both SeFa and WarpedSpace have the highest correlation with ``identity", implying that their variations of these attributes are often coupled with variations of the face identity during the traversal. By contrast, our method has the best attribute correlations mostly on the diagonal, which explicitly indicates that these attributes of our method are more disentangled from each other.

\begin{table}[h]
    \centering
    \caption{Comparison of the VP scores (\%) with pre-trained VAEs. The results are averaged over $3$ random runs.}
    \resizebox{0.89\linewidth}{!}{
    \begin{tabular}{c|c|c|>{\columncolor[gray]{0.8}}c}
    \toprule
         \rowcolor{white}\textbf{Models} & \textbf{WarpedSpace} & \textbf{Ours (Linear)} &\textbf{Ours}\\
    \midrule
          \textbf{MNIST}  &13.44 & 12.76 &\textbf{17.38} \\
          \textbf{dSprites} &15.01 & 14.25 & \textbf{18.49} \\
    \bottomrule
    \end{tabular}
    }
    \label{tab:vp_score_vae}
\end{table}

\subsection{Results with Pre-trained VAEs}

\begin{figure}[htbp]
    \centering
    \includegraphics[width=0.99\linewidth]{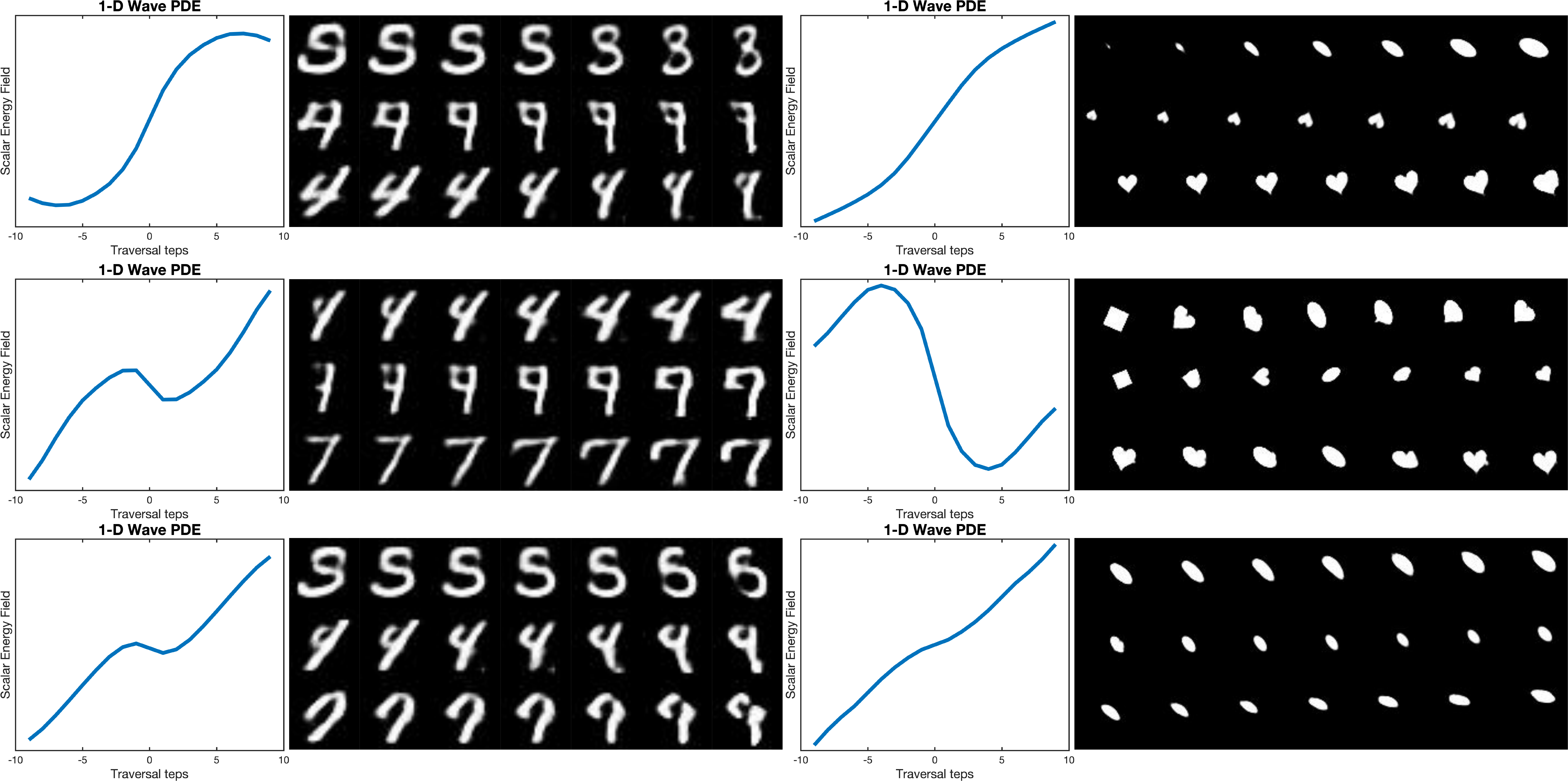}
    \caption{Exemplary semantic attributes and the corresponding traversal trajectories with VAEs trained on MNIST and dSprites.}
    \label{fig:vae_pretrained}
\end{figure}

Fig.~\ref{fig:vae_pretrained} displays the exemplary semantics discovered by our method with pre-trained VAEs. Our potential PDEs exhibit a diverse set of different shapes and the interpolation images correspond to distinct transformation factors. Table~\ref{tab:vp_score_vae} presents the quantitative evaluation of VP scores. The linear baseline and WarpedSpace achieve similar performance, falling behind our method by $4\%$. This demonstrates again the effectiveness of our PDEs in modelling latent traversal.

\begin{table}[htbp]
    \centering
    \caption{The log-likelihood $\log p_{\theta}(\vx)$ evaluated over the dataset.}
   \resizebox{0.99\linewidth}{!}{
    \begin{tabular}{c|c|>{\columncolor[gray]{0.8}}c}
    \toprule
        \rowcolor{white}\textbf{Models} &  \textbf{Naively Trained} &  \textbf{Trained with Our Method}\\
    \midrule
        \textbf{MNIST} & -2207.70 & \textbf{-2144.71} \\
        \textbf{dSprites} & -3848.04 & \textbf{-3740.97}\\
    \bottomrule
    \end{tabular}
    }
    \label{tab:log-likelihood}
\end{table}

\subsection{Results with VAEs Trained from Scratch}
\label{sec:supervised_exp}

\begin{figure}[h]
    \centering
    \includegraphics[width=0.99\linewidth]{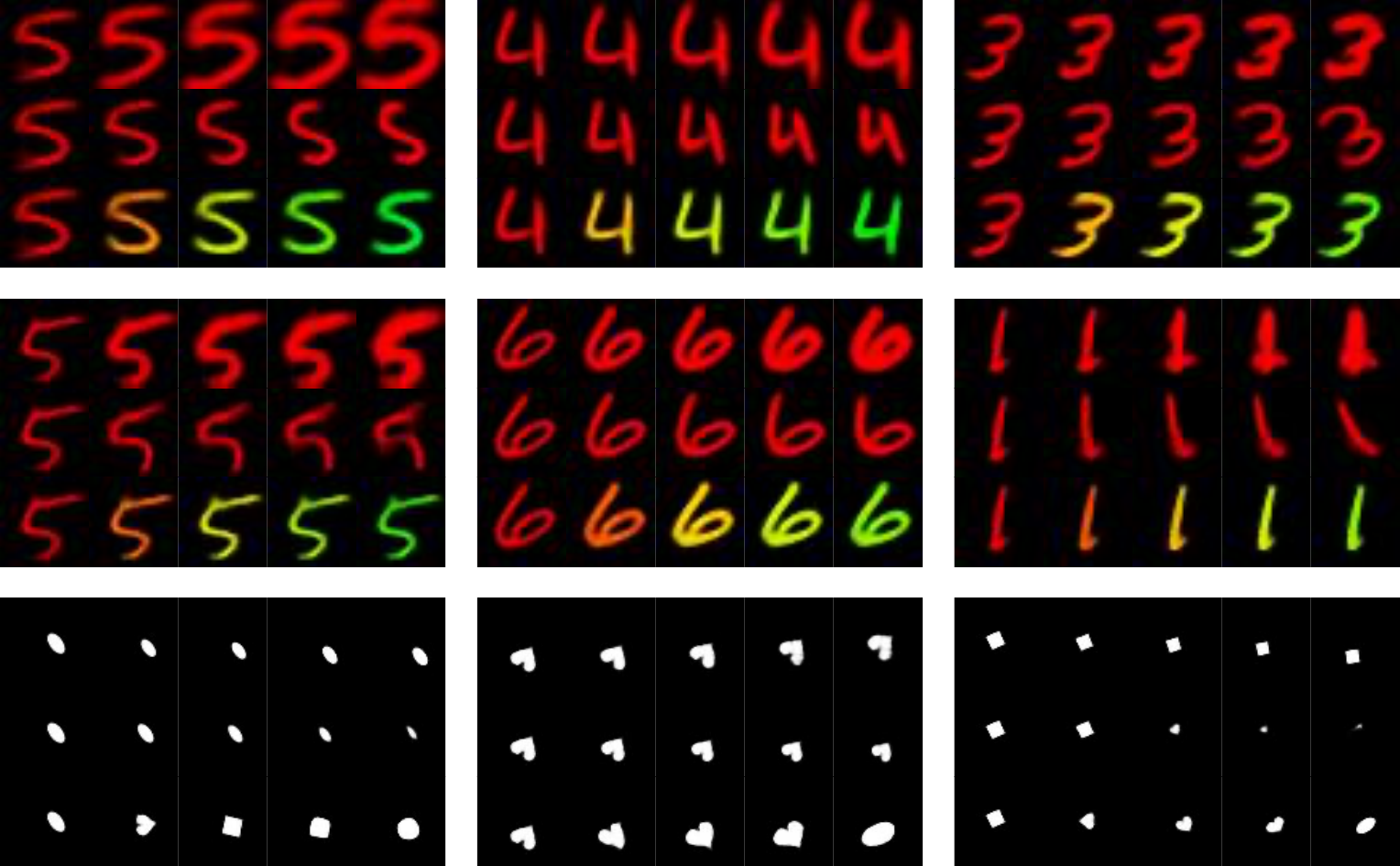}
    \caption{Exemplary traversal results when our method is integrated into the VAE training process. For MNIST, the exhibited transformations are scaling, rotation, and coloring changes from top to bottom. For Dsrpites, the corresponding transformations are y-axis position, scaling, and shape changes from top to bottom.}
    \label{fig:vae_scratch}
\end{figure}

Table~\ref{tab:log-likelihood} compares the log-likelihood of VAEs integrated with our method. Notice that common disentanglement methods would often sacrifice the likelihood~\citep{higgins2016beta}. However, integrating our PDEs into the training process slightly improves the likelihood estimation. Fig.~\ref{fig:vae_scratch} displays the exemplary traversal results of the pre-defined transformations. Our method is also able to learn and generalize the pre-defined transformation factors well. 

\begin{table}[h]
    \centering
    \caption{Equivariance error on MNIST.}
    \resizebox{0.89\linewidth}{!}{
    \begin{tabular}{c|c|c|c}
    \toprule
         \rowcolor{white}\textbf{Transformations} & \textbf{Rotation} & \textbf{Scaling} & \textbf{Coloring}  \\
    \midrule
          \textbf{Our Method} & \cellcolor{gray!40}{\textbf{235.96}} & \cellcolor{gray!40}{\textbf{230.39}} & \cellcolor{gray!40}{\textbf{240.64}}\\
         \textbf{Vanilla VAE} &1278.21& 1309.56&1370.54 \\
    \bottomrule
    \end{tabular}
    }
    \label{tab:equivariance}
\end{table}

One interesting geometric property induced by our potential flows is the approximate equivariance for VAEs trained from scratch. At a high level, an equivariant map is one which commutes with a desired transformation group, \emph{i.e.,} $T'[f(x)]=f(T[x])$. This can be understood as preserving geometric symmetries of the input space. The gradient of our potential function can be interpreted as the equivariant latent operator $T'$ corresponding to the observed input transformation $T[x]$. As is typical in the equivariance literature, we can measure how close this is to exact equivariance by measuring the equivariance error:
\begin{equation}
\begin{aligned}
  \texttt{Err} &= \sum_{t=1}^{T} | \vx_{t} -\hat{\vx}_{t} |\\
    &= \sum_{t=1}^{T}| \vx_{t} - \texttt{Decode}(\vz_0 + \sum^{t}\nabla_\vz u^k) |
\end{aligned}
\end{equation}
We see this is equivalent to measuring the satisfaction of the equivariance relation $T[x]-f^{-1}(T'[f(x)])=0$ where $f^{-1}$ is approximated with the decoder. Table~\ref{tab:equivariance} presents the evaluation results against a vanilla VAE on transforming MNIST. Note that since the vanilla VAE has no notion of a corresponding transformation in the latent space $T'$ (\emph{i.e.,} no a priori known latent structure), we simply set $\nabla_\vz u^k$ to $0$ and treat this as a lower bound baseline. We see that our method performs significantly above this baseline, indicating that it could be helpful to build equivariant VAEs. 

%In future work, we would be very interested to see how this compares with more competitive baselines for learned equiviariance.

\subsection{Discussions}

%\noindent\textbf{Gallery of Discovered Attributes.}

\noindent\textbf{Linear Directions as Special Cases.} We note that the linear traversal approaches can be understood as special cases of our second-order wave equations. Actually, for general linear functions defined as $u(x,t) = a\cdot x + b\cdot t$ where $a$ and $b$ denote the coefficients, the solutions would all correspond to wave equations. In this sense, linear functions are simplified special cases of our waves. One piece of evidence for supporting this is that in certain cases where the structure of the latent space might be simple, our PDEs can also reduce back to functions that are almost linear, such as the traversal paths of the semantic attribute ``Eye Size" in Fig.~\ref{fig:quali_anime_dog} and the transformation of scaling in Fig.~\ref{fig:vae_pretrained} right.

\begin{figure}[h]
    \centering
    \includegraphics[width=0.99\linewidth]{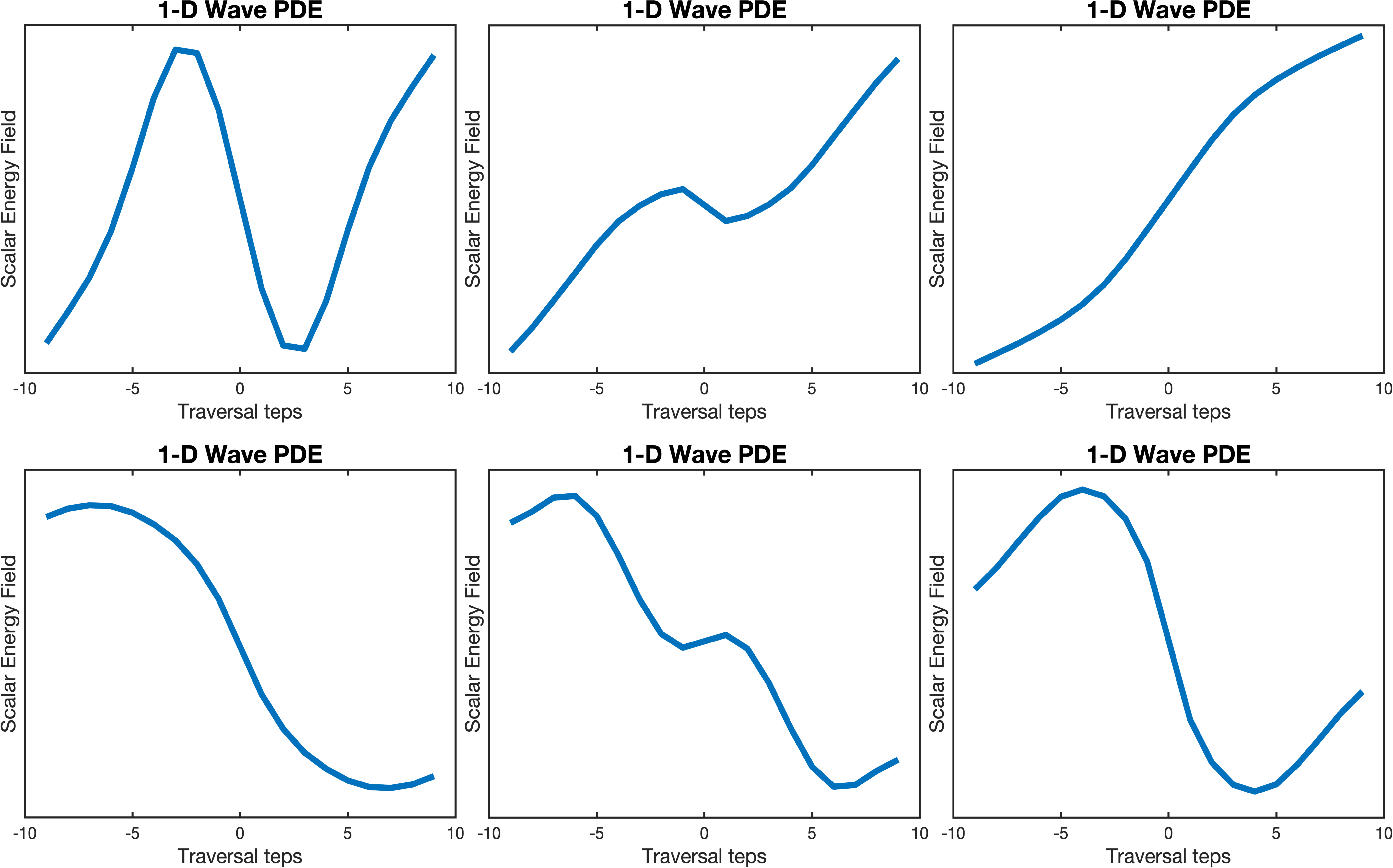}
    \caption{Common shapes of potential PDEs in our experiments.}
    \label{fig:heat_type}
\end{figure}

\noindent\textbf{Path Diversity.} Our potential PDEs can be very different in shape and period. Fig.~\ref{fig:heat_type} exhibits some common PDEs learned in our experiments. As can be seen, our wave equations allow for a wide set of traversal paths, ranging from linear lines to traveling waves of a full period. This flexibility enables modeling diverse trajectories in the manifold.

\begin{figure}[htbp]
    \centering
    \includegraphics[width=0.99\linewidth]{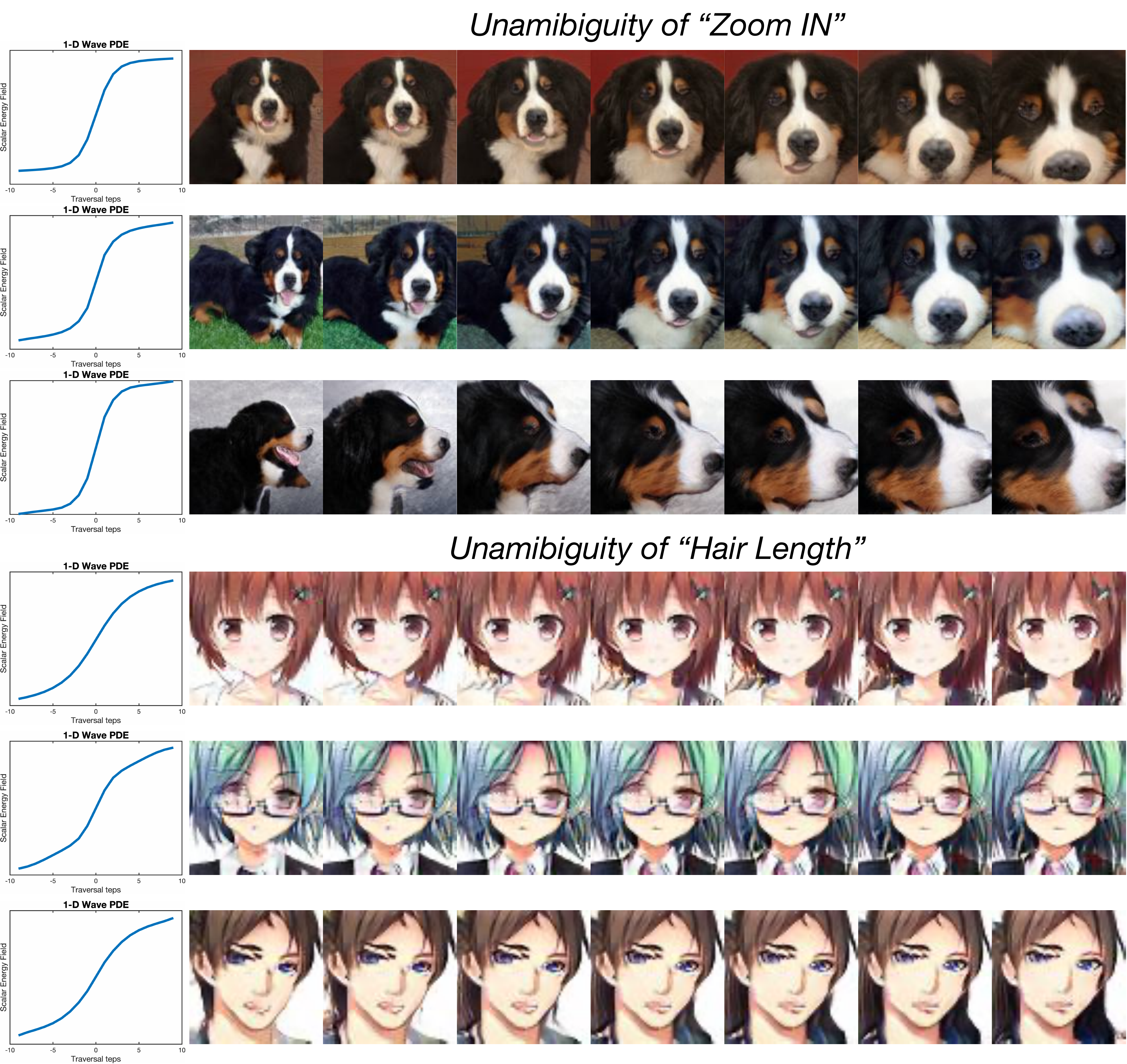}
    \caption{Unambiguity of our potential PDEs and the corresponding discovered semantics: the shape of trajectory and the image attribute of a traversal path are consistent to different samples.}
    \label{fig:unambiguity}
\end{figure}

\noindent\textbf{Semantic and Trajectory Unambiguity.} As shown in Fig.~\ref{fig:unambiguity}, for the same traversal path, the semantic attribute is consistent to different samples and the corresponding PDE paths are of very similar shapes. Take the semantic attribute of ``Zoom IN" as an example. The scalar potential energy fields of the three images all have slow changes near the endpoints while taking sharp increases in the middle regime. Accordingly, the interpolation images coincide with identical semantics.

%Driven by our learned velocity field $\nabla u(\vz,t)$, the probability distribution $\rho(\vz,t)$ defined over $\gZ$ space and time could satisfy certain evolution PDEs. In particular, with certain $\nabla u(\vz,t)$, the evolution of $\rho(\vz,t)$ could possibly become some special PDEs, such as heat equations, Fokker Planck equations, and Porous Medium equations. The specific types depend on the modeled relation between $\nabla u(\vz,t)$ and $\rho(\vz,t)$. For instance, if the velocity field learns $\nabla u(\vz,t) = - \nabla \log(\rho(\vz,t))$, the evolution of $\rho(\vz,t)$ would become heat equations. More details about the possible relations are kindly referred to~\citet{santambrogio2017euclidean}.

%The specific types depend on the modeled relation between the velocity field $\nabla u(\vz,t)$ and the probability distribution $\rho(\vz,t)$. We derive their connection and discuss the possible special PDEs of $\rho(\vz,t)$ in the supplementary material. 

%\noindent\textbf{Novel Perspective of Disentanglement.} We take a new perspective to set up disentanglement and latent traversal. 

%\noindent\textbf{Local Editing Applications.} 

%\subsection{Ablation Studies}

%\noindent\textbf{Impact of Jacobian Regularization.}

%\noindent\textbf{Impact of timestep.}

\noindent\textbf{Geometric Properties of Latent Spaces.} Besides the equivariance property of the encoder/decoder, we also have some novel observations about the shape and variations of $\nabla_{\vz}u^k$. For VAEs, we observe that the \textbf{\textit{simple}} variation factors that involve \textbf{\textit{linear}} transformations (\emph{e.g.}, scaling and translation shown in Fig.~\ref{fig:vae_pretrained} right) tend to be accordingly \textbf{\textit{more linear}} in the latent space. For GANs, the semantic attributes that edit \textbf{\textit{local}} image regions tend to be \textbf{\textit{more linear}} in the latent space, such as the attribute ``Eye Size" in Fig.~\ref{fig:quali_anime_dog} and the attributes ``Glasses" and ``Hat Length" in Fig.~\ref{fig:quali_stylegan2}. Through all the experiments, the traversal directions generally tend to have fewer variations when closer to the endpoints. We think this is because at the endpoints (\emph{i.e.}, large timesteps) our potentials learnt to not violate the semantic attribute and not to go out of the data manifold.

\noindent\textbf{Limitations of Potential Flows.} It is known that potential flows are limited in their ability to represent all forms of physically known flows. For example, since the curl of the gradient is known to be zero, potential flows are inherently irrotational and thus cannot model vorticity. In the case of latent traversals, the literature largely appears to model non-cyclic transformations (such as hair length or skin color), and thus this modeling assumption is observed to be valid. However, this limitation explains why the rotation traversals attempted to be learned by our VAE model perform poorly. Ultimately, we propose this framework as a first step towards modeling latent traversals with more complex, physically informed dynamics, and suggest that in some settings, these physical biases may match the underlying data in a beneficial way. We propose that valuable future work could explore alternative parameterizations of the latent vector field which could respectively yield alternative biases suitable to other datasets. 

\noindent\textbf{Alternative PDE Modeling Approaches.} We mainly explore the PINN-based physical constraints to model our PDEs. Despite the flexibility and efficiency, this approach achieves the \textit{soft} PDE constraints approximately. Other alternative possibilities for PDE modeling include Neural Conservation Laws~\citep{richter2022neural} that impose \textit{hard} divergence-free constraints and accurate neural PDE solvers~\citep{hsieh2019learning,brandstetter2022message}. Investigating other PDE modeling approaches is an important research direction in future work.

\noindent\textbf{Famous PDEs of the Sample Evolution.} Driven by our learned velocity field $\nabla u(\vz,t)$, the sample evolution of $\vz$ over space and time could satisfy certain PDEs. In particular, with certain $\nabla u(\vz,t)$, the evolution of $\vz$ could possibly become some special well-known PDEs, such as heat equations, Fokker Planck equations, and Porous Medium equations. The specific types depend on the relation between $\nabla u(\vz,t)$ and $\rho(\vz,t)$. For instance, if the velocity field is set as $\nabla u(\vz,t) = - \nabla \log(\rho(\vz,t))$, the evolution would become the heat equations. More details about the possible relations are kindly referred to~\citet{santambrogio2017euclidean}. 
\section{Conclusion}

%In this paper, we rethink the set-up of latent traversal from the paradigm of optimal transport theory and its fluid mechanics interpretation. We model a set of disentangled traversal paths that correspond to the shortest distance between distributions with learned PDEs as Wassserstein gradient flows.

Inspired by the fluid mechanical interpretation of optimal transport and the role of traveling waves in neuroscience, we propose to model the latent traversal flexibly by the gradient flows of learned dynamic potential landscapes. Our method can model a set of traversal paths with distinct semantics to improve the disentanglement ability of pre-trained GANs and VAEs. Furthermore, our PDEs can be integrated into the training process of VAEs as regularization on the latent space to improve the model likelihood estimation. 

\section*{Acknowledgements}

This research was supported by the EU H2020 project AI4Media (No. 951911). Yue Song acknowledges travel support from ELISE (GA no 951847). Andy Keller thanks the Bosch Center for Artificial Intelligence for funding.

\bibliography{main}
\bibliographystyle{icml2023}

%%%%%%%%%%%%%%%%%%%%%%%%%%%%%%%%%%%%%%%%%%%%%%%%%%%%%%%%%%%%%%%%%%%%%%%%%%%%%%%
%%%%%%%%%%%%%%%%%%%%%%%%%%%%%%%%%%%%%%%%%%%%%%%%%%%%%%%%%%%%%%%%%%%%%%%%%%%%%%%
% APPENDIX
%%%%%%%%%%%%%%%%%%%%%%%%%%%%%%%%%%%%%%%%%%%%%%%%%%%%%%%%%%%%%%%%%%%%%%%%%%%%%%%
%%%%%%%%%%%%%%%%%%%%%%%%%%%%%%%%%%%%%%%%%%%%%%%%%%%%%%%%%%%%%%%%%%%%%%%%%%%%%%%
\newpage
\appendix
\onecolumn
\section{Appendix}

\subsection{Implementation Details}

\noindent\textbf{VP Score.} The dataset of image pairs $[\vx_{0},\vx_{T}]$ is created by randomly sampling from different interpretable directions. Since the used models have a different number of directions, the crafted datasets also have a different number of images accordingly. Specifically, the dataset consists of $10,000$ images for SNGAN and VAEs, $20,000$ images for BigGAN, and $40,000$ images for StyleGAN2. We randomly select $10\%$ of the images as the training set and the rest as the test set. The simple neural network for the VP score evaluation consists of four stacked convolutional layers with batch normalization and ReLU activations. The learning rate is set to $0.005$, and we train the network for $300$ epochs with the batch size set as $32$. We report the classification accuracy (\%) on the test set as the score.

\noindent\textbf{Pre-trained GANs.} We set the total timestep $T$ to $10$ for all the datasets and models. In line with~\citet{Tzelepis_2021_ICCV}, the number of potential functions (traversal paths) $K$ is set as $64$ for SNGAN, $120$ for BigGAN, and $200$ for StyleGAN2. The output images are of size $64{\times}64$ for SNGAN, of size $256{\times}256$ for BigGAN, and of size $1024{\times}1024$ for StyleGAN2. During the inference stage, we also negatively traverse the latent space by $\vz_t - \nabla_\vz u^k(\vz_t, t)$. The anti-symmetry of the traversal is thus achieved.

%\noindent\te\vztbf{SNGAN with AnimeFace.}

%\noindent\te\vztbf{BigGAN with ImageNet.}

%\noindent\te\vztbf{StyleGAN2 with FFHQ.}

\noindent\textbf{Pre-trained VAEs.} We set the number of traversal path $K$ to $32$ and define the total timestep $T$ as $10$ for both MNIST and Dsprites. The training process lasts $100,000$ iterations. 

%\noindent\te\vztbf{Pre-trained VAE with MNIST.}

%\noindent\te\vztbf{Pre-trained VAE with Dsprites.}
\noindent\textbf{Integrating Traversal into VAE Training.} For MNIST, we define $3$ factors of variations, \emph{i.e.,} scaling, rotation, and color transformations. Each transformation has $8$ states of variations. For Dsprites, we use the self-contained $5$ factors of variations, \emph{i.e.,} x position, y position, scaling, orientation, and shape transformations. The training also lasts $100,000$ iterations for both datasets. For the comparison fairness, the naively trained baseline employs the loss $\E_{\vz_t}[ {-}\log p_{\theta}(\vx_t | \vz_t) {+} \mathrm{D}_{\text{KL}}\left[q_{\phi}(\vz_t | \vx_t) || p_{\gZ}(\vz_t) \right ]$ to optimize the ELBO of the same transformed input data.

\subsection{Impact of Different Losses}

\begin{table}[h]
    \centering
    \caption{Impact of different loss terms on the VP scores (\%).}
    %\resizebox{0.79\linewidth}{!}{
    \begin{tabular}{c|c|c|c|c}
    \toprule
         \rowcolor{white}\textbf{Models} & \textbf{Ours} & \textbf{w/o $\gL_{J}$} & \textbf{w/o $\gL_{f}$} & \textbf{w/o $\gL_{u}$}  \\
    \midrule
          \textbf{SNGAN}  &\textbf{65.89} & 55.37 & 45.78 & 63.19 \\
          \textbf{BigGAN} &\textbf{15.29} &13.91 &12.87 &14.68 \\
          \textbf{StyleGAN2} &\textbf{48.54} & 41.77 & 36.91 &46.24 \\
    \bottomrule
    \end{tabular}
    %}
    \label{tab:abla}
\end{table}

Table~\ref{tab:abla} presents the complete ablation studies of losses on all the datasets. As can be seen above, when $\gL_{J}$, $\gL_{f}$, or $\gL_{u}$ are not applied, our model would have performance degradation of different extents. The Jacobian regularization $\gL_{J}$ can encourage that the trajectory could cause meaningful variations, while the PDE constraints $\gL_{f}$ ensures that the potential flow follows wave-like spatial-temporal dynamics. The initial condition constraint can improve the score slightly but more importantly it is applied to help generate smoother traversal paths.

\subsection{Why We Need PDE Constraints}

We add the PDE constraints to the velocity fields to learn good spatial-temporal dynamics for smooth, continuous, and flexible latent trajectories.  The formulation matches the space dynamics $\nabla u$ to the time dynamics 
$\partial_t u$, leading to stable potential flows and smooth wave-like paths in the latent space. Since the latent code is progressively updated by $\vz_{t+1} = \vz_{t} + \nabla_\vz u^k(\vz_t, t)$, if no constraints are applied on the gradient, the magnitude of $\nabla_\vz u^k(\vz_t, t)$ might gradually get amplified and then eventually the latent code $\vz$ is likely to go out of the manifold. Enforcing PDE constraints in spatiotemporal form could help to limit the magnitude of the gradient and create wave-like plausible trajectories. %This is because the potential dynamics in time $\frac{\partial}{\partial t} u^k(\vz_{t},t)$ is relatively more stable as the timestep update $t=t+1$ is fixed at every step.

\subsection{Visual Gallery of Identified Semantic Attributes}

\begin{figure*}[h]
    \centering
    \includegraphics[width=0.8\linewidth]{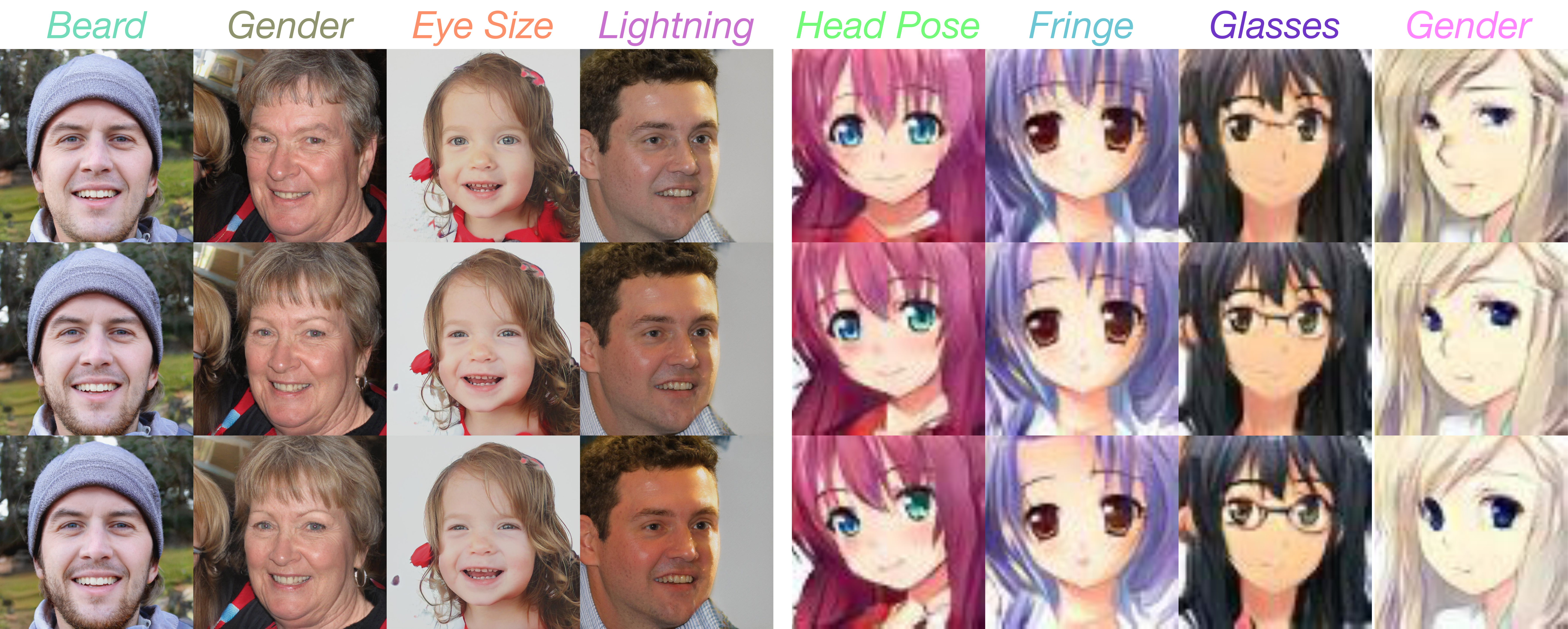}
    \caption{More semantics discovered by our learned potential PDEs on SNGAN and StyleGAN2.}
    \label{fig:stylegan_anime_app}
\end{figure*}

\noindent\textbf{SNGAN and StyleGAN2.} Fig.~\ref{fig:stylegan_anime_app} displays some more semantic attributes identified by our potential PDEs on SNGAN and StyleGAN2. Our method can precisely control the target image attributes while keeping other traits uninfluenced.

\begin{figure*}[h]
    \centering
    \includegraphics[width=0.7\linewidth]{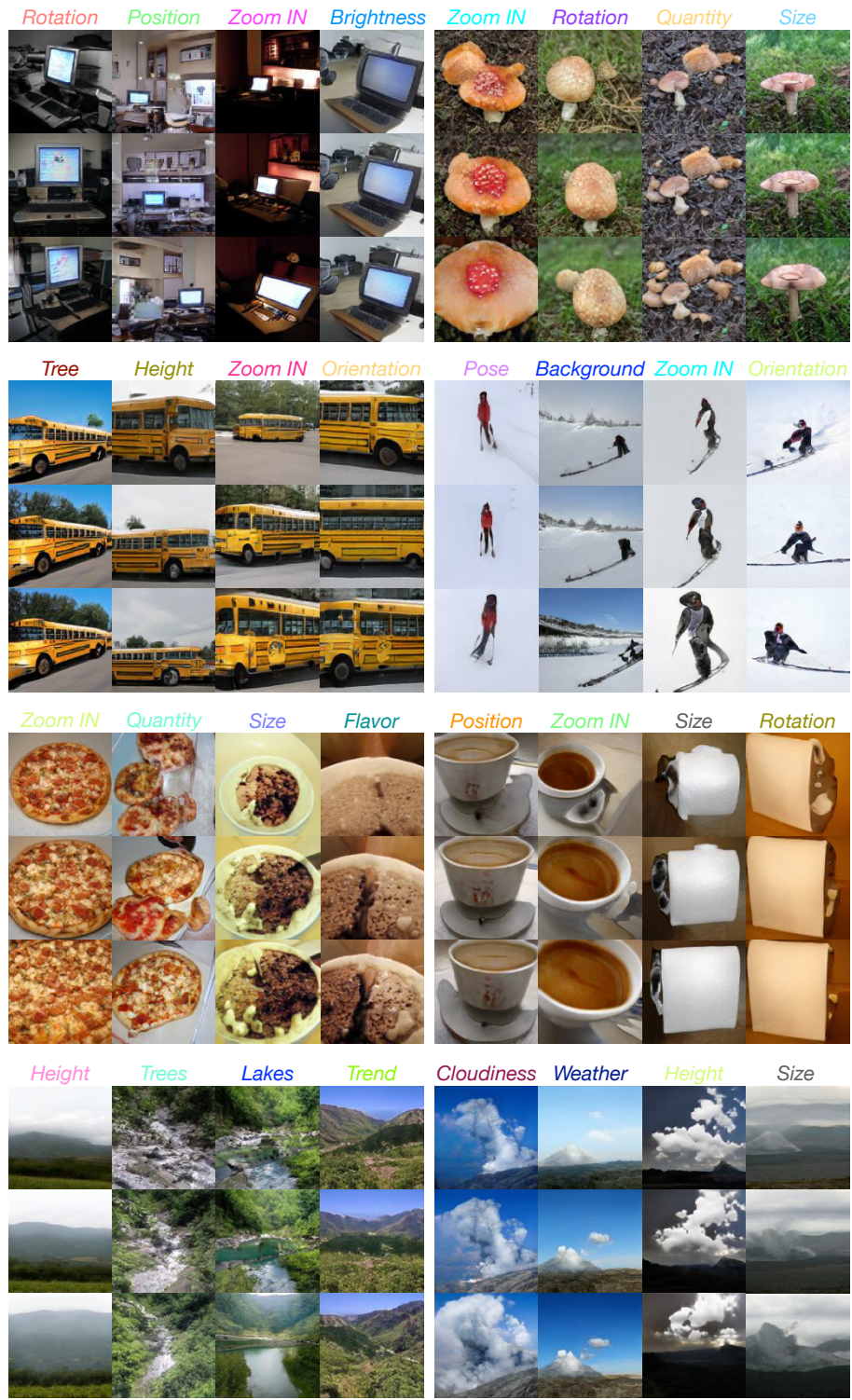}
    \caption{Semantic attributes of different objects discovered by our learned potential PDEs on BigGAN. The specific image categories include Computer Screen ($1_{st}$ row left), Mushroom ($1_{st}$ row right), Schoolbus ($2_{nd}$ row left), Ski ($2_{nd}$ row right), Pizza and Ice Cream ($3_{rd}$ row left), Coffee and Toilet Tissue ($3_{rd}$ row right), Valley ($4_{th}$ row left), and Volcano ($4_{th}$ row right).}
    \label{fig:biggan_app}
\end{figure*}

\noindent\textbf{BigGAN.} Previous disentanglement approaches heavily rely on human faces and animal images for visualization. Here we instead show some results with alternative objects belonging to the ImageNet classes based on BigGAN. Fig.~\ref{fig:biggan_app} presents such traversal results. Our potential PDEs are still able to identify distinct semantics from images of various categories.

\end{document}